# An Image Processing Pipeline for Camera Trap Time-Lapse Recordings

Michael L. Hilton, Mark T. Yamane, Leah M. Knezevich
Eckerd College
St. Petersburg, Florida, USA
{hiltonm, mtyamane, lmknezev}@eckerd.edu

## Abstract

*A new open-source image processing pipeline for analyzing camera trap time-lapse recordings is described. This pipeline includes machine learning models to assist human-in-the-loop video segmentation and animal re-identification. We present some performance results and observations on the utility of this pipeline after using it in a year-long project studying the spatial ecology and social behavior of the gopher tortoise.*

## 1. Introduction

Machine learning (ML) has recently captured the interest of wildlife researchers, who hope it will be able to automatically analyze the vast amounts of data collected by camera traps. Initial efforts have shown ML is a promising tool for that [1], but for most researchers, ML is still a laboratory curiosity that has not made its way into routine use as part of their field studies. Many questions need to be answered before ML is broadly adopted by the wildlife community, ranging from the practical (such as how to train, execute, and maintain ML systems), to the methodological (e.g., how the accuracy of ML systems affects study conclusions), and fundamental concerns (e.g., will investment in this technology result in new knowledge about my study species, or just more digits of precision for things I already know?). To explore these questions and others, we created an open-source software pipeline for processing time-lapse camera trap images as part of an ongoing study monitoring spatial ecology and social behavior of the gopher tortoise (*Gopherus polyphemus*).

Gopher tortoises are large, long-lived terrestrial turtles endemic to the Southeastern Coastal Plains ecoregion of North America [2]. They are listed at the state or federal level as threatened or endangered species throughout most of their range. Gopher tortoises spend most of their time in and around excavated burrows which are typically 3–6 m long and 2 m deep. Tortoise burrows provide a resting place and shelter from predators and fire for over 300 different species, making them a keystone species in their ecosystem [3]. Gopher tortoises routinely interact with each other, forming social networks that contain cliques [4]. We believe a better understanding of gopher tortoise social networks will be useful in the management of tortoise populations, and obtaining this improved understanding is a main driver of the work described here.

A mounded area of bare soil or sand, called the apron, is often found at the tortoise burrow entrance. The apron is the focal point for many tortoise behaviors, such as basking, mating, nesting, and competitive interactions. Given the importance of burrows and aprons to gopher tortoise activities, they are frequently the focus of camera trap-based research [5,6]. To learn more about tortoise social interactions, we recorded time-lapse camera trap images at the aprons of 12 occupied burrows in the Boyd Hill Nature Preserve, located in St. Petersburg, Florida USA, during the period of November 2020 through October 2021.

To analyze the 130 TB of image data collected, we turned to ML. After evaluating the publicly available camera trap software, we found no systems that met our needs. The existing software was unsatisfactory for one or more of the following reasons: the system's animal detector does not include our species of interest; the system is cloud-based, and we cannot afford the networking bandwidth required to transmit our massive data sets in a reasonable period of time; or the system is not designed specifically for time-lapse recordings, so its user interface and work processes are not streamlined for our needs. To overcome these deficits, we created a suite of software tools and automated work processes specifically to create, segment, and annotate camera trap time-lapse recordings with a minimum of user effort. This paper describes our software pipeline and the results we have obtained so far. Although the examples presented here involve tortoises, our software also works with ML detectors trained for other species.

## 2. Image Processing Pipeline

The job of the image processing pipeline is to transform time-lapse images taken by camera traps into compressed video files accompanied by data identifying the video

CVIP 2022 Computer Vision for Animals Workshop (CV4Animals), June 20th, 2022

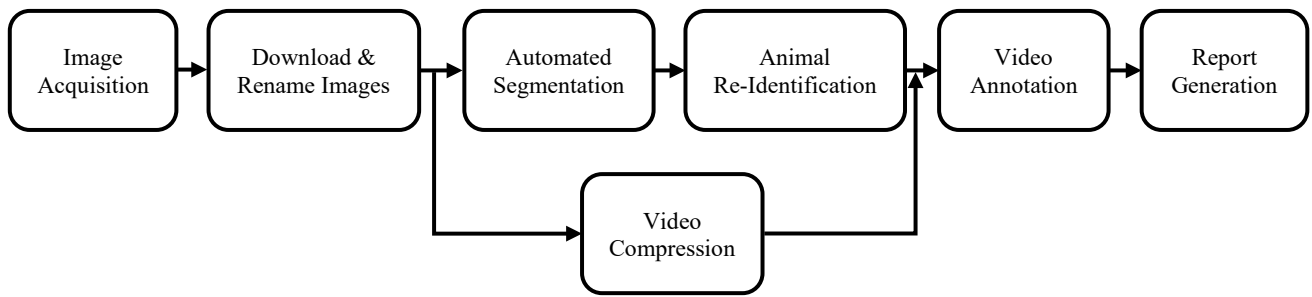

Figure 1: Logical diagram of the pipeline used to analyze time-lapse videos of gopher tortoise burrow entrances.

segments containing gopher tortoises and predictions about the identities of those tortoises. Figure 1 is a logical diagram of the image processing pipeline; in the actual implementation, some of the modules are combined into a single program and other modules are composed of multiple programs. Each module is written in Python and described in more detail below.

*Image Acquisition*: Two trail cameras (SL122 Pro, Meidase, Shenzhen, China) are deployed per burrow, one mounted directly above the burrow entrance to aid in the identification of tortoise individuals using their carapace markings, and the other placed ~6 m in front of the entrance to provide a wide field of view for observing social interactions occurring in the vicinity of the burrow. Each camera is assigned a unique serial number encoding its burrow and viewpoint. Figure 2 shows an example of the overhead images recorded in this study. Each camera acquires one 4-megapixel image (2048 × 1536 pixels) every 5 s during daylight hours (07:00–20:00).

*Download & Rename Images*: Camera trap SD memory cards are collected several times each week for processing. Nine SD cards are loaded at one time into a bank of SD card readers. A custom program copies images from the SD cards to a hard drive. As part of the copy process, each image is given a unique name composed from its burrow identifier, camera view, and acquisition date and time. To automate the copy process, this information is extracted using optical character recognition of the text in the information banner burned into the bottom of each image; the operator provides no direct input to the program – they just load the SD cards into the readers and run the program.

*Video Compression*: For long-term storage, each day's time-lapse images are compressed into one MP4 video file per camera. Compression is performed using ffmpeg, a popular open-source video processing tool (ffmpeg.org). In addition to creating individual video files for each camera day, a composite video is created for each burrow day showing a time-aligned side-by-side view of the burrow's overhead and frontal cameras.

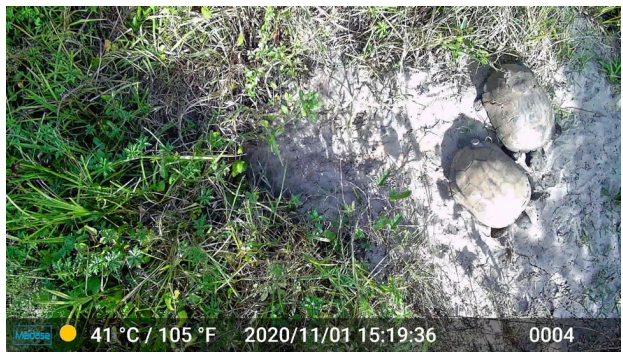

Figure 2: Example of the overhead images taken at a burrow entrance. The overhead view facilitates identifying individual tortoises using their carapace shape and coloration.

*Automated Video Segmentation*: Each image is examined by an EfficientDet-based object detector [7] implemented using the Tensorflow Object Detection API [8] trained on 550 tortoise images using the fine-tuning transfer learning technique [9]. A draft video segmentation is created by grouping together consecutive image frames found to contain one or more tortoises.

*Animal Re-Identification*: A Siamese neural network (SNN) is used for animal re-identification [10]. Our SNN backbone is an Xception network from the Tensorflow Keras library, which takes a 299×299-pixel image as input. The SNN has 32 embedding dimensions and was trained using transfer learning and a triplet loss function [11]. The training set consisted of 10 images each of 26 unique tortoises.

It has been suggested that keeping objects in the same orientation can improve SNN training and re-identification accuracy by keeping features in the same regions of the images [12]. To this end, we create vertically oriented "mugshots" of tortoises by using Mask R-CNN [13] to extract the outline of each tortoise detected in an image, fitting an ellipse to that outline, and rotating the image so the major axis of the ellipse is vertical. The SNN was trained using mugshots of tortoises in the head up orientation. Since the tortoise in a query mugshot could be either head up or head down, the re-identification algorithm tries both possibilities, picking the top five predictions obtained from running the SNN twice – once on a mugshot and again on its 180° rotation.

Table 1: Confusion matrix comparing the video segmentation performance of three human graders and the machine learning segmentation algorithm. True positives are grader segments with at least one image that overlaps a ground truth segment. False negatives are ground truth segments with no overlapping grader segment. False positives are grader segments that do not overlap a ground truth segment.

|  | True positives | False positives | False negatives | Accuracy | Precision | Recall |
|---|---|---|---|---|---|---|
| Grader 1 | 135 | 5 | 10 | 96% | 96% | 93% |
| Grader 2 | 135 | 5 | 10 | 96% | 96% | 93% |
| Grader 3 | 138 | 2 | 7 | 97% | 99% | 95% |
| Segmentation Algorithm | 130 | 56 | 15 | 79% | 70% | 90% |

The SNN is run on a small number of image frames from each video segment previously identified as containing tortoises, generating a top-5 prediction for each tortoise in a frame. These predictions are displayed by the manual video annotation module to assist humans in identifying tortoises.

*Video Annotation*: The draft video segmentations can be reviewed and edited using the video annotation module. This program includes frame-level navigation, a timeline display for viewing and navigating annotation segments, and annotation editing tools. The editing tools can be configured via a text file to annotate arbitrarily named events, which can be associated with specified animal IDs if desired. The program displays the top-5 re-identification results for the current frame to aid in selecting the correct animal ID.

*Report Generation*: A variety of reports can be generated for tracking logistical matters, such as the status of image capture and processing, and video annotation data. The annotation data reports are comma-separated value files suitable for analysis by Excel or statistical packages such as R.

## 3. Performance

### 3.1. Speed of Execution

The pipeline presented in this paper is capable of supporting projects that are continuously producing large numbers of images – more images than can be economically stored in their native format. To prevent a backlog of image data, the download, segmentation, and compression modules of the pipeline must be highly reliable and execute faster than the rate at which new images are acquired.

Each day, our 24 camera traps capture ~250,000 images requiring ~375 GB of disk storage. We are using 32MB SD cards in our cameras, which can store two 14-hour days' worth of 4-megapixel images taken at a rate of one image every 5 s. We process this data on a single Dell Alienware Aurora R8 desktop computer (32 GB memory, 8 Intel i7 cores, 3.60 GHz) equipped with a GPU (NVIDIA GeForce RTX 2080, 8 GB memory). ML video segmentation is performed on images taken by the overhead camera at a rate of ten images/s, allowing a 32MB SD card containing 20,000 images to be processed in just under 34 minutes. Images taken by the front view camera do not need to be segmented, so downloading and renaming occurs at the rate of 30 images/s, allowing a 32MB SD card to be processed in 11 minutes. The Video Compression module requires a total of 34 minutes to generate the three MP4 videos created for each burrow day. Altogether, processing two days' worth of images requires ~23 hours, well below the 48-hour deadline needed to prevent a backlog of images. As a practical matter, beating the 48-hour deadline requires a human operator to reload our 9-bay SD card reader twice a day.

### 3.2. Video Segmentation

The performance of the automated video segmentation module was measured on time-lapse recordings taken at eight tortoise burrows during a pilot study from 25 November 2020 to 30 November 2020. All time-lapse recordings were segmented by the automated segmentation module and manually by three human graders. Two graders were students with minimal segmentation experience and the third grader was the lead author, who is an experienced segmenter. The graders are henceforth referred to as Graders 1, 2, and 3, respectively. Manual segmentation was performed using the video annotation module using a combination of recording playback and scrubbing, where the grader drags the playback slider across the timeline to quickly review a recording. Each grader worked in isolation at their own pace. After all graders completed segmenting the recordings, Grader 3 created a best approximation of the ground truth by considering the segmentations produced by all three human graders and the automated algorithm while slowly and carefully re-segmenting the recordings.

Early in our project it became evident that fully automated segmentation was unlikely given our resources, so our goal has been to use ML to generate a draft segmentation that assists a human in quickly creating a highly accurate segmentation. If the automated algorithm correctly identifies at least one image containing tortoises per ground truth segment, thereby drawing the human's attention to that region of the recording, the detector has properly done its job. The performance of the segmentation algorithm is presented in Table 1 as a confusion matrix for the human and machine graders summarizing the number of ground truth segments that overlap with at least one image in a grader's segment (true positives), the number of ground truth segments with no grader segment overlap (false negatives), and the number of grader segments that do not overlap a ground truth segment (false positives).

Tortoises appeared in a total of 145 segments of the test dataset. All three human graders outperformed the animal detector with regards to segmentation accuracy (Table 1). At a minimum required detection confidence of 0.90, the segmentation algorithm's recall performance approaches to within 3 percentage points of student Graders 1 and 2, with roughly one false positive per recording and one false negative per four recordings.

### 3.3. Re-Identification

The animal re-identification module was tested on a set of 100 unseen tortoise images (5 images per tortoise, 20 individuals that were all part of the training set). The module achieved a top 1 accuracy of 77% and a top 5 accuracy of 89%. Performance was improved to a top 1 accuracy of 89% and top 5 accuracy of 95% by pruning the module's reference library, which originally contained 10 images per tortoise, down to a single image per tortoise. The pruning was accomplished by randomized removal of images whose absence did not lower the top 1 accuracy score.

## 4. Discussion

We began this paper by listing a few of the "big questions" wildlife researchers have about machine learning (indeed, about any new technology). We are endeavouring to answer some of these questions by incorporating ML-based workflows into the routine work processes of an on-going camera trap study of gopher tortoises.

We believe that computer-savvy researchers can reapply our pipeline to their own studies, and we have made our software freely available on (https://github.com/hiltonml/camera_trap_tools.git). The pipeline can be used with or without ML models. Full reapplication will require training ML models for the species of interest. The software download includes a Google Colab notebook for training an image classification-based animal detector that can be used in the video segmentation module. Implementing an animal re-identifier is more complicated and likely requires assistance from an ML expert.

After a year's experience using our pipeline, we believe it has great utility, processing large volumes of data with minimal human effort. Without the high level of automation the pipeline provides, it would not have been feasible for one researcher to continuously monitor a dozen tortoise burrows for a full year while devoting only 2 to 3 hours a week to data management. Most of that time was spent manually verifying/editing the automated video segmentations for each week's overhead recordings (~11.5 hours of viewing if played at 30 frames/s).

The only reason we performed a manual verification of each video segmentation is because we are creating a year-long ground truth data set for use in benchmarking new ML algorithms. In a different application context, one might be able to dispense with some or all manual verification. For example, in our tortoise study, we are primarily interested in social interactions between tortoises. In our years' worth of data, the segmentation algorithm only missed 15 of the 984 (1.5%) social interactions present. It is doubtful the missed interactions would change any conclusions reached regarding tortoise behavior.